\documentclass{article}

\usepackage{microtype}
\usepackage{graphicx}
\usepackage{subfigure}
\usepackage{capt-of}
\usepackage{placeins}
\usepackage{booktabs} 

\usepackage{hyperref}

\usepackage[preprint]{icml2026}

\usepackage{amsmath}
\usepackage{amssymb}
\usepackage{mathtools}
\usepackage{amsthm}

\usepackage[capitalize,noabbrev]{cleveref}

\theoremstyle{plain}

\newcommand{\R}{\mathbb{R}}

\newcommand{\cE}{\mathcal{E}}
\newcommand{\cF}{\mathcal{F}}

\icmltitlerunning{YuriiFormer}

\begin{document}
\twocolumn[
\icmltitle{{YuriiFormer}: A Suite of Nesterov-Accelerated Transformers}

\begin{icmlauthorlist}
\icmlauthor{Aleksandr Zimin}{MIT}
\icmlauthor{Yury Polyanskiy}{MIT}
\icmlauthor{Philippe Rigollet}{MIT}
\end{icmlauthorlist}

\icmlaffiliation{MIT}{MIT}

\icmlcorrespondingauthor{P. Rigollet}{rigollet@mit.edu}

\icmlkeywords{Transformers, Optimization, Acceleration, Dynamical Systems}

\vskip 0.3in
]
\printAffiliationsAndNotice{} 


\begin{abstract}
We propose a variational framework that interprets transformer layers as iterations of an optimization algorithm acting on token embeddings. In this view, self-attention implements a gradient step of an interaction energy, while MLP layers correspond to gradient updates of a potential energy. Standard GPT-style transformers emerge as vanilla gradient descent on the resulting composite objective, implemented via Lie--Trotter splitting between these two energy functionals. This perspective enables principled architectural design using classical optimization ideas. As a proof of concept, we introduce a Nesterov-style accelerated transformer that preserves the same attention and MLP oracles. The resulting architecture consistently outperforms a nanoGPT baseline on TinyStories and OpenWebText, demonstrating that optimization-theoretic insights can translate into practical gains.
\end{abstract}



\section{Introduction}

Transformers dominate modern sequence modeling, but their architecture is still largely an empirical design. Attention, MLPs, residual connections, and normalization are known to be essential, yet their combined effect is rarely viewed as a coherent algorithm. As a result, architectural changes are mostly heuristic, and principled methods for modifying transformer blocks remain limited.

This work builds on two complementary lines of research. First, neural architectures can be derived from numerical schemes for continuous-time dynamics, where alternative discretizations or operator splittings naturally yield new architectures. \citet{LuLiHe19} obtain transformer variants from splitting schemes for ODEs. This line of work shows that classical numerical analysis can guide design rather than ad hoc choices. Second, attention mechanisms admit a variational interpretation: viewing tokens as interacting particles, self-attention can be seen as a preconditioned gradient step of an interaction energy~\cite{geshkovski2025mathematical, sander2022sinkformers}, with close connections to Wasserstein gradient flows~\cite{ambrosio2005gradient,CheNilRig25}, mean-field dynamics~\cite{chen2025quantitative,rigollet2025meanfield}, and synchronization phenomena such as the Kuramoto model~\cite{acebron2005kuramoto,criscitiello2024synchronization,andrew25}.

We unify these ideas by viewing \emph{transformers as optimization algorithms on token configurations}. Each layer is a discrete step of an optimization method on an implicit objective over all token embeddings. Attention layers act as first-order oracle calls to the gradient of an \emph{interaction energy} encoding token--token interactions, while MLP layers query a \emph{potential energy} acting independently on each token; depth corresponds to iteration count.

Under this interpretation, standard GPT-style transformer blocks that alternate attention and MLP layers implement vanilla gradient descent on the composite objective via Lie--Trotter splitting. Viewing transformers through this optimization lens opens the architecture to a broad class of alternative optimization and splitting schemes. As a concrete example, we replace vanilla gradient descent with a Nesterov-style accelerated method while preserving the same attention and MLP oracle structure. The resulting architecture, \textsc{YuriiFormer}, consistently improves over a nanoGPT baseline on TinyStories and OpenWebText, demonstrating that this perspective leads to effective and principled architectural modifications.

\section{Transformers as optimizers}
\label{sec:composite_optimization}

Transformer architectures update collections of token embeddings through two fundamental mechanisms: attention layers, which mix information across tokens, and MLP layers, which apply nonlinear transformations independently to each token.  
In this section we show that these mechanisms admit a unified variational interpretation.  
Specifically, attention and MLP layers implement gradient-based updates of two complementary energy functionals defined over token configurations: an \emph{interaction energy} encoding token--token interactions, and a \emph{potential energy} acting independently on each token.

From this perspective, a transformer block realizes a concrete first-order optimization procedure for a composite objective given by the sum of these two energies.  
Classical GPT-style architectures correspond to a particular choice of splitting scheme for this composite optimization problem, a viewpoint that will later allow us to systematically modify transformer blocks using ideas from numerical optimization.

\subsection{Attention as a gradient of an interaction energy}

Self-attention defines an explicit pairwise interaction among tokens.  
Given a token configuration $X := (x_1,\ldots,x_n) \in (\mathbb{R}^d)^n$, a single attention layer with a residual connection updates each token according to\footnote{Additional components such as normalization, multi-head structure, and causal masking are standard in practice and used in our experiments; since they are not required for the conceptual developments here, we defer them to Section~\ref{sec:experiments}.}
\begin{equation}
\label{eq:att}
x_i \;\leftarrow\; x_i \;+\; 
\frac{\sum_{j=1}^n V_t x_j \, e^{\langle Q_t x_i, K_t x_j\rangle}}
     {\sum_{j=1}^n e^{\langle Q_t x_i, K_t x_j\rangle}},
\end{equation}
where $Q_t$, $K_t$, and $V_t$ denote the query, key, and value matrices at layer~$t$.

The structure of~\eqref{eq:att} places transformers within the class of interacting particle systems~\cite{LuLiHe19,dutta2021redesigning}.
Building on this perspective, \citet{geshkovski2025mathematical} showed that self-attention can be interpreted as the gradient flow of an \emph{interaction energy} defined over the space of token configurations, providing a precise variational description of attention dynamics.
This variational characterization serves as the foundation for several studies of how attention organizes and propagates representations across layers~\cite{sander2022sinkformers,cowsik2024geometric,rigollet2025meanfield}, and it is also the starting point of the present work.

Concretely, consider the interaction energy
\[
\mathcal{E}(X)
\;:=\;
\sum_{i,j=1}^n e^{\langle x_i, x_j\rangle}.
\]
With respect to a suitable Riemannian metric~\cite{geshkovski2025mathematical,andrew25}, the gradient of $\mathcal{E}$ with respect to each token takes the form
\[
\nabla_{x_i}\mathcal{E}(X)
\;=\;
\frac{\sum_{j=1}^n x_j\, e^{\langle x_i, x_j\rangle}}
     {\sum_{j=1}^n e^{\langle x_i, x_j\rangle}},
\qquad i=1,\dots,n.
\]
Comparing this expression with~\eqref{eq:att} shows that an attention layer implements a gradient step on the interaction energy, modulated by two learnable transformations:
\[
\setlength{\arraycolsep}{3pt}
\begin{array}{lrl}
\text{(i) Preconditioning:}
& \nabla_{x_i}\mathcal{E} &\longmapsto V_t\,\nabla_{x_i}\mathcal{E}, \\[0.6em]
\text{(ii) Change of coordinates:}
& \langle x_i,x_j\rangle &\longmapsto \langle Q_tx_i, K_tx_j\rangle.
\end{array}
\]

At the level of the full token configuration $X$, the attention layer therefore realizes an update of the form
\[
X \;\leftarrow\; X + {\sf Attn}_t(X),
\]
where ${\sf Attn}_t(X)$ is a modulated gradient update of the interaction energy $\mathcal{E}$.

\subsection{MLPs as a gradient of a potential energy}

In contrast to attention, an MLP layer acts independently on each token.  
Its residual update is given by
\begin{equation}
\label{eq:mlp}
x_i \;\leftarrow\; x_i \;+\; W_t\,\sigma(A_t x_i + b_t),
\qquad i=1,\dots,n,
\end{equation}
where $A_t$ and $W_t$ are weight matrices, $b_t$ is a bias vector, and the nonlinearity $\sigma$ is applied entry-wise.

This update also admits a variational interpretation.  
Consider the potential energy
\[
\mathcal{F}(X)
\;:=\;
\sum_{i=1}^n V(x_i),
\qquad
V(x) := \sum_{\ell=1}^d v(x^{(\ell)}),
\]
where $v:\mathbb{R}\to\mathbb{R}$ is a scalar function and $x=(x^{(1)}, \ldots, x^{(d)})$ with derivative $v'=\sigma$, then $\nabla V(x) = \sigma(x)$ where as before, $\sigma$ is applied entry-wise to $x$. 

Under this identification, the MLP update corresponds to a gradient step on the potential energy $\mathcal{F}$, modulated by a layer-dependent preconditioner $W_t$ and an affine change of coordinates $x \mapsto A_t x + b_t$.  
At the configuration level, we write this update as
\[
X \;\leftarrow\; X + {\sf MLP}_t(X),
\]
emphasizing its role as a gradient-based transformation acting independently on each token.
\subsection{Composite optimization}

Attention and MLP layers therefore implement complementary gradient updates of an interaction energy $\mathcal{E}$ and a potential energy $\mathcal{F}$.  
Their composition within a transformer block realizes a concrete first-order algorithm for optimizing the composite objective $\mathcal{E} + \mathcal{F}$ over token configurations.

Previous work~\cite{geshkovski2025mathematical} interprets simplified attention dynamics (with $V_t = Q_t = K_t = I_d$) as gradient ascent on the interaction energy, leading to clustering behavior.  
In practice, however, the learned value matrix $V_t$ can induce attractive, repulsive, or mixed effects, effectively modulating the sign and geometry of the interaction~\cite{geshkovski2024emergence,bruno2025multiscale}.  
As different layers may realize qualitatively different dynamics, we do not view transformers as consistently minimizing or maximizing a fixed energy.  
Rather, we adopt the broader perspective that transformer blocks implement a structured \emph{optimization procedure} for a composite objective, without committing to a global sign convention.

From this viewpoint, architectural design amounts to choosing a first-order method for composite optimization.  
A central role in such methods is played by \emph{splitting schemes}, which arise naturally from time discretizations of differential equations of the form
\begin{equation}
\label{eq:ode1}
\dot X(t) = {\sf Attn}_t(X(t)) + {\sf MLP}_t(X(t)).
\end{equation}
A forward Euler discretization yields the parallel update
\begin{equation}
\label{eq:gd}
X \;\leftarrow\; X + {\sf Attn}_t(X) + {\sf MLP}_t(X),
\end{equation}
which is implemented in transformer architectures such as PaLM~\cite{palm}.  

Modern transformers typically favor \emph{Lie--Trotter splitting}, in which attention and MLP updates are applied sequentially:
\begin{equation}
\label{eq:gdlt}
\begin{array}{cc}
X &\leftarrow\; X + {\sf Attn}_t(X),\\
X &\leftarrow\; X + {\sf MLP}_t(X).
\end{array}
\end{equation}
More systematic explorations of alternative splitting schemes were carried out in~\cite{LuLiHe19}, where Strang--Marchuk splitting leads to modified residual streams and moderate but consistent empirical improvements.  
These results support the view that transformer architecture design can be framed as the selection of a numerical scheme for composite optimization.

\section{Nesterov-Accelerated Transformers}

The update rules~\eqref{eq:gd} and~\eqref{eq:gdlt} can be interpreted as instances of gradient descent on the composite energy $\mathcal{E}+\mathcal{F}$, once attention and MLP layers are viewed as modulated gradient oracles for the interaction and potential energies, respectively.  
Moreover, the characteristic alternation between attention and MLP layers in standard transformer blocks is not imposed by gradient descent itself, but arises from the choice of a Lie--Trotter splitting scheme for this composite objective.  
Under this interpretation, a transformer block corresponds to a single iteration of a first-order optimization method, with the oracle structure determined by attention and MLP layers and the block structure determined by the splitting scheme.

With this viewpoint in place, architectural design can be decoupled into two choices: a \emph{first-order optimization template} and a \emph{splitting scheme} for its implementation.  
In this section, we replace the gradient descent template with a \emph{Nesterov accelerated gradient template}, while preserving the same interaction and potential energy oracles.  
This substitution leads to a family of accelerated transformer architectures, collectively referred to as \textsc{YuriiFormer}.  
Different splitting schemes give rise to different architectural variants; for concreteness, we focus on a Lie--Trotter realization, which exhibits strong practical performance.

\subsection{Acceleration and momentum-based optimization}

Nesterov’s accelerated gradient (NAG) augments gradient descent with a momentum variable that propagates information across iterations~\cite{nesterov1983method,nesterov2004introductory}.  It was proposed in the early eighties by Yurii Nesterov to achieve the optimal iteration complexity $O(1/t^2)$ on smooth convex objectives after the realization that the pervasive Gradient Descent (GD) algorithm was suboptimal.

Given an objective function $f$, NAG maintains a state $x_t$ and a velocity $v_t$ and proceeds via the following three steps:
\begin{equation}
\label{eq:nag}
\begin{aligned}
\text{(Lookahead)} \qquad
x_{t+\frac12} &= x_t + \mu_t v_t, \\[0.3em]
\text{(Velocity update)} \qquad
v_{t+1} &= \beta_t v_t - \gamma_t \nabla f(x_{t+\frac12}), \\[0.3em]
\text{(State update)} \qquad
x_{t+1} &= x_t + v_{t+1},
\end{aligned}
\end{equation}
where $\mu_t, \beta_t$ and $\gamma_t$ are explicit deterministic sequences.  

A defining feature of Nesterov acceleration is the evaluation of the gradient at the \emph{lookahead point} $x_{t+\frac12}$ rather than at the current state $x_t$. For comparison, Polyak’s heavy-ball method~\cite{Polyak1964} omits the lookahead step and evaluates the gradient at the current iterate,
\[
v_{t+1} = \beta_t v_t - \gamma_t \nabla f(x_t),
\qquad
x_{t+1} = x_t + v_{t+1}.
\]
This corresponds to taking $\mu_t=0$ in NAG.

Adaptive methods such as Adam~\cite{kingma2015adam} further introduce coordinate-wise rescaling of the gradient that can be folded into the learned preconditioning matrix.  Indeed, in the framework developed here, attention and MLP layers already implement learned preconditioning and changes of variables through their parameterization.  
Consequently, classical momentum schemes collapse into a single \emph{acceleration template} when expressed at the level of modulated gradient oracles.  
We therefore adopt Nesterov acceleration as a representative and analytically clean choice for introducing momentum into transformer architectures, without modifying the underlying attention or MLP oracles.

\subsection{YuriiFormer architectures}

Using the notation of Section~\ref{sec:composite_optimization}, we replace gradient-oracle calls to $\nabla f(\cdot)$ in~\eqref{eq:nag} with ${\sf Attn}_t(X)$ and ${\sf MLP}_t(X)$ acting on \emph{token velocity} $V=(v_1,\ldots,v_n) \in (\R^d)^n$ associated to the token configuration $X=(x_1,\ldots,x_n)$ resulting into two dependent streams.

We now describe two instantiations of \textsc{YuriiFormer}, corresponding to Euler discretization and Lie--Trotter splitting respectively.

\paragraph{YuriiFormer with Euler discretization.}

\begin{figure}[t]
  \centering
  \includegraphics[width=0.25\textwidth,height=0.35\textheight,keepaspectratio]{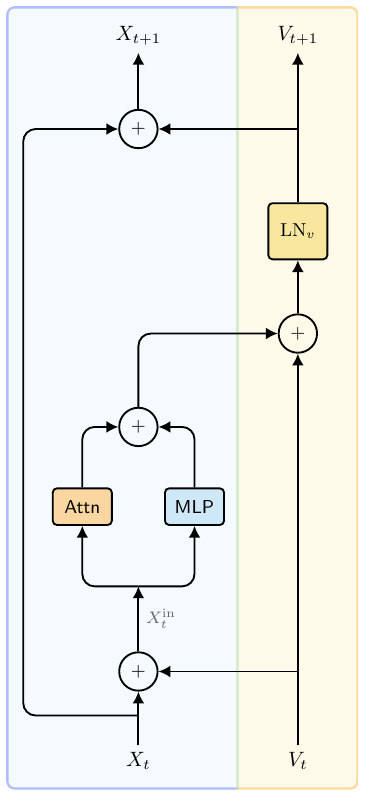}\hspace{2em}
  \includegraphics[width=0.23\textwidth,height=0.35\textheight,keepaspectratio]{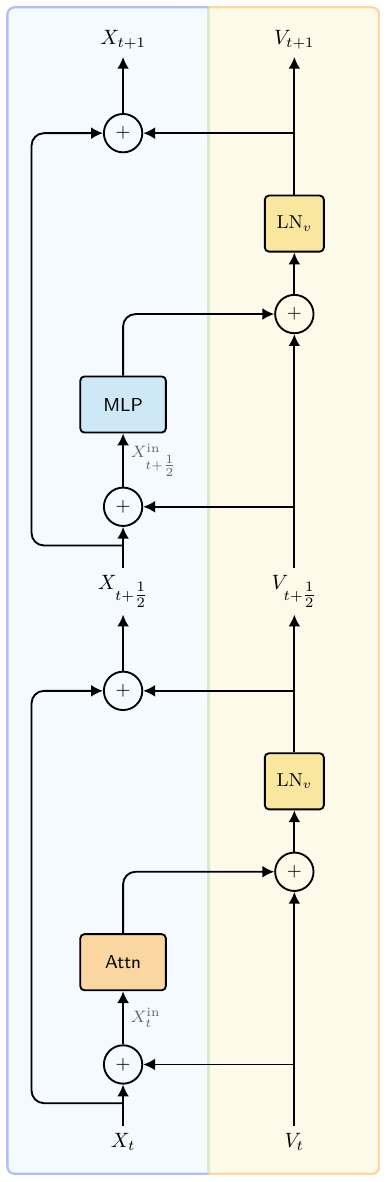}
  \caption{{\sc YuriiFormer} pre-layerNorm and Euler discretization (left) and Lie--Trotter splitting (right), with velocity LayerNorm $\mathrm{LN}_v$ applied after each velocity update.}
  \label{fig:arch-no-splitting}
\end{figure}

The three steps of NAG in this context take the form
\[
\begin{aligned}
X^{\mathrm{in}}_t &= X_t + \mu_t V_t, \\
V_{t+1} &= \beta_t V_t + \gamma_t {\sf Attn}_t(X^{\mathrm{in}}_t) + \gamma_t {\sf MLP}_t(X^{\mathrm{in}}_t), \\
X_{t+1} &= X_t + V_{t+1},
\end{aligned}
\]
where $\mu_t,\beta_t\in(0,1)$ are momentum parameters and $\gamma_t>0$ is a step size.  
A schematic representation of this architecture enhanced with pre-layerNorm is shown in Figure~\ref{fig:arch-no-splitting} (left).

\paragraph{YuriiFormer with Lie--Trotter splitting.}
Alternatively, acceleration on a composite objective can be combined with Lie--Trotter splitting to obtain a standard sequential composition of attention and MLP layers:
\[
\begin{aligned}
X^{\mathrm{in}}_t &= X_t + \mu_t V_t, \\
V_{t+\frac12} &= \beta_t V_t + \gamma_t {\sf Attn}_t(X^{\mathrm{in}}_t), \\
X_{t+\frac12} &= X_t + V_{t+\frac12}, \\
X^{\mathrm{in}}_{t+\frac12} &= X_{t+\frac12} + \mu_{t+\frac12} V_{t+\frac12}, \\
V_{t+1} &= \beta_{t+\frac12} V_{t+\frac12} + \gamma_{t+\frac12} {\sf MLP}_t(X^{\mathrm{in}}_{t+\frac12}), \\
X_{t+1} &= X_{t+\frac12} + V_{t+1}.
\end{aligned}
\]

This variant mirrors the structure of modern GPT-style transformers while injecting momentum at the representation level.  
The resulting architecture is illustrated in Figure~\ref{fig:arch-no-splitting} (right).

\medskip

Both variants preserve the same optimization template and differ only in the splitting scheme used to combine oracle calls across layers. In Section~\ref{sec:extensions}, we also consider Polyak’s heavy-ball method (no lookahead).
In Section~\ref{sec:experiments}, we evaluate these architectures empirically and show that Nesterov-style acceleration yields consistent improvements over non-accelerated baselines.

\section{Related work}
Our work sits at the intersection of dynamical-systems views of deep networks, variational analyses of attention, and optimization-based interpretations of transformers.

\paragraph{Dynamical systems and splitting views.}
Interpreting deep networks as discretizations of continuous-time dynamics goes back to viewing residual networks as ODE/PDE solvers and optimal control problems~\citep{weinan2017proposal,haber2017stable,chen2018neural}. Within this perspective, transformer architectures have been modeled as multi-particle dynamical systems, where attention and feed-forward layers arise as diffusion and convection terms in a convection--diffusion equation, and standard transformer blocks correspond to Lie--Trotter splitting schemes~\citep{LuLiHe19}. Alternative splitting schemes and depth-wise evolution operators have been used to redesign transformer layers and improve parameter efficiency~\citep{dutta2021redesigning}. Momentum-based dynamics have also been used to design new neural architectures; see e.g., \cite{rusch2021coupled} for RNNs and~\cite{momentumresnets} for ResNets, as well as \citet{wang2022momentum} for a synthesis. Unlike these previous lines of work, we focus explicitly on viewing standard GPT-style blocks as discrete-time optimization algorithms for a composite objective.

\paragraph{Variational and energy-based views of attention.}
Several works endow attention with an explicit variational or energy-based structure, often through interacting-particle models of token dynamics and their mean-field limits~\citep{bruno2025emergence,bruno2025multiscale,chen2025critical,chen2025quantitative,chen2025clustering,criscitiello2024synchronization,geshkovski2024dynamic,geshkovski2024emergence,geshkovski2025mathematical,karagodin2025normalization,andrew25}. In a related line, energy-based formulations design explicit energies over token configurations and view the forward pass as (discretized) gradient descent on these energies~\citep{hoover2023energytransformer}. Our interaction-energy viewpoint is broadly related to this line of work, but we pair it with a complementary potential energy for MLPs to obtain a composite objective tailored to full GPT-style transformer blocks.

\paragraph{Transformers as optimization algorithms.}
A complementary line of work reconstructs transformer layers from energy-minimization principles. \citet{yang2022transformers} show that one can construct an energy function whose gradient-descent iterations closely match the transformer forward pass, providing an unfolded optimization perspective. More recently, and closest to our work, \citet{ren2025intrinsic} introduce an energy-based framework with local energies \(E_i\), a global energy \(F\), and an optimization algorithm, and obtain both existing and new attention mechanisms (including momentum-, Nesterov-, and Newton-style variants) as one-step optimization updates. We instead embed Nesterov-style acceleration directly into the transformer block as a two-stream (state--velocity) architecture acting on token configurations, while preserving the usual attention and MLP oracles and the Lie--Trotter block structure used in GPT-style models.


\section{Experiments}
\label{sec:experiments}
To demonstrate the effectiveness of different instantiations of \textsc{YuriiFormer},
we compare against a nanoGPT baseline~\cite{karpathy2022nanogpt},
trained with the Muon optimizer~\cite{jordan2024muon} on two datasets:
TinyStories~\cite{eldan2023tinystories} and OpenWebText~\cite{gokaslan2019openwebtext}.

\subsection{Setup}

\paragraph{Models.}
We use decoder-only Transformer language models trained autoregressively with a causal attention mask.
All models add a learned position embedding to each token embedding, use a context length of 1024, and apply a final LayerNorm before a weight-tied output projection.

Each layer uses a pre-normalization layout with two LayerNorms: one applied to the residual stream before the attention sublayer and one applied before the feedforward network (FFN/MLP).
Self-attention uses $h$ heads with embedding dimension $d$ (head dimension $d/h$), a fused QKV projection, causal scaled dot-product attention, and an output projection.
The FFN is a two-layer MLP with GELU and a 4$\times$ expansion (hidden size $4d$).

The baseline updates the residual stream via sequential additions of the attention and FFN outputs, while \textsc{YuriiFormer} variants keep the same attention/MLP modules but modify the depth-update rule.
For variants that maintain an additional velocity state (e.g., Nesterov and Polyak's heavy ball),
we apply a dedicated LayerNorm to the velocity after each velocity update (after each substep in split updates).
We train two Transformer configurations\footnote{We use the nomenclature $L/H/d$ to denote a Transformer configuration with $L$ layers, $H$ heads and embedding dimension~$d$.}: $12L/12H/768d$ {\tt (small)} on TinyStories and OpenWebText and $24L/16H/1024d$ {\tt (medium)} on OpenWebText.
Using a standard convention both the GD baselines and {\sc YuriiFormer} variants come at 124M parameters for {\tt small} and 354M parameters for {\tt medium}.

\paragraph{Training.}
All models are trained with Muon using a mixed Muon+AdamW optimizer.
Across all runs we use AdamW $(\beta_1,\beta_2)=(0.9,0.95)$, Muon momentum $\beta=0.95$, and global gradient-norm clipping at 1.0. We train in \texttt{bfloat16} and use a cosine learning-rate schedule with linear warmup to the peak learning rates in Table~\ref{tab:training-hparams}; the minimum learning rate is $0.1\times$ the peak.
Table~\ref{tab:training-hparams} summarizes the main training hyperparameters.

\begin{table}[ht]
  \caption{Training budgets and peak learning rates for TinyStories (TS) and OpenWebText (OWT). Train tokens = steps $\times$ tokens/step.}
  \label{tab:training-hparams}
  \begin{center}
    \begin{small}
      \begin{sc}
        \begin{tabular}{lccc}
          \toprule
          & TS 12L & OWT 12L & OWT 24L \\
          \midrule
          Model size & 123.6M & 123.6M & 353.6M \\
          Train steps & 10k & 30k & 30k \\
          Train tokens & 4.92B & 14.75B & 14.75B \\
          Warmup steps & 1k & 3k & 3k \\
          Muon LR & 0.02 & 0.004 & 0.006 \\
          AdamW LR & 6e-4 & 6e-4 & 6e-4 \\
          \bottomrule
        \end{tabular}
      \end{sc}
    \end{small}
  \end{center}
  \vskip -0.1in
\end{table}

Additional experimental protocol details (datasets/preprocessing, optimizer parameter groups, and evaluation/checkpointing) are provided in Appendix~\ref{app:exp-setup}.

We use a global batch size of 30 sequences with 16-step gradient accumulation at context length $T{=}1024$ (491{,}520 tokens per optimizer step), which corresponds to about 960 steps per TinyStories epoch and 17.5k steps per OpenWebText epoch.
Overall, we train for about 10.4 epochs on TinyStories and about 1.7 epochs on OpenWebText.
For reference, most main runs were executed on 2$\times$ NVIDIA H200 GPUs (some auxiliary runs used other GPUs); we did not optimize throughput.



\subsection{Cross-entropy loss}

We compare the baseline and \textsc{YuriiFormer} variants under identical training budgets.
For each dataset and model size, all methods use the same batch size, sequence length, and number of optimizer steps, and are trained on the same deterministic sequence of training batches.
This controls for data-order randomness, so differences between loss curves primarily reflect the update rule.
We evaluate next-token prediction using validation cross-entropy loss (nats/token; lower is better).

\begin{figure*}[t]
  \centering
  \subfigure[\textbf{Training loss.}]{\includegraphics[width=0.48\textwidth]{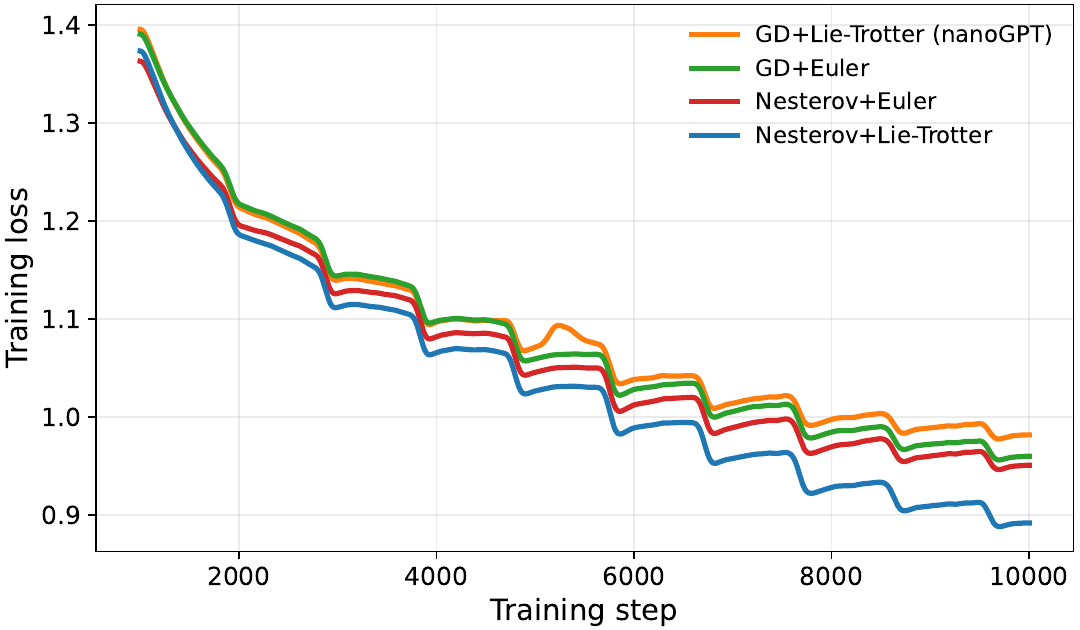}}
  \hfill
  \subfigure[\textbf{Validation loss.}]{\includegraphics[width=0.48\textwidth]{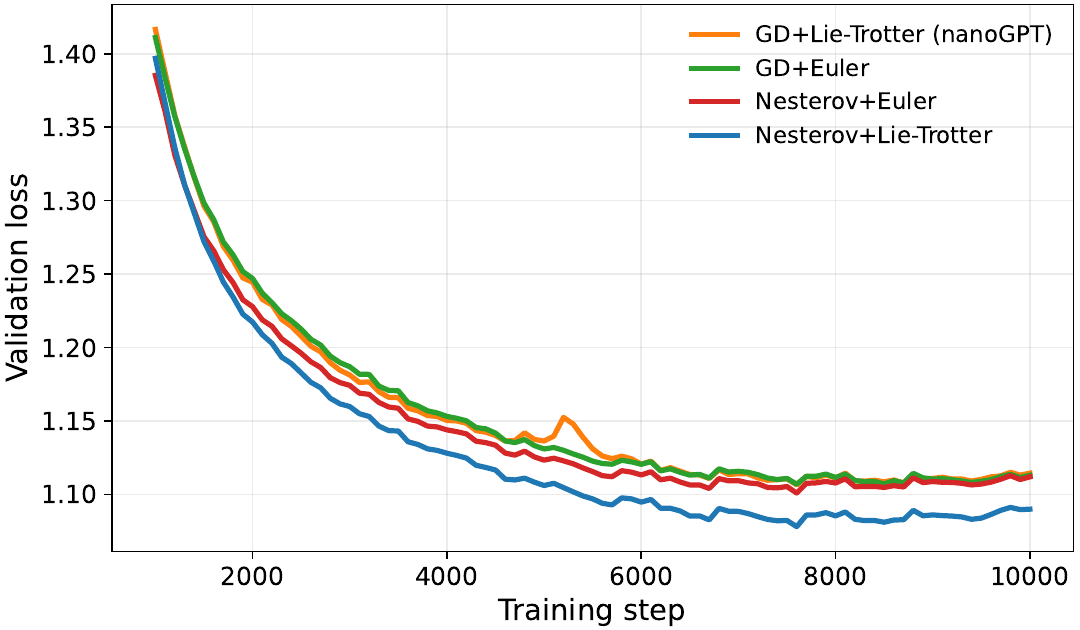}}
  \caption{TinyStories (\texttt{small}): loss vs.\ training step (steps $\ge 1000$). Training loss is Gaussian-smoothed ($\sigma=50$ steps) for readability.}
  \label{fig:tinystories-loss-curves}
\end{figure*}

\paragraph{TinyStories ({\tt small}; 10k steps).}
As shown in Figure~\ref{fig:tinystories-loss-curves}, with about 10 epochs of training, validation loss reaches its minimum around step 7.6k for all methods and changes little thereafter, while the training loss continues to decrease, indicating overfitting late in training.
Nesterov+Lie--Trotter achieves the lowest validation loss; Table~\ref{tab:tinystories-losses} lists best and final losses.

\begin{table}[ht]
  \caption{TinyStories ({\tt small}; 10k steps): next-token prediction losses (nats/token; lower is better). Best = min validation loss; Final = validation loss at 10k steps; Train @10k = training loss at 10k.}
  \label{tab:tinystories-losses}
  \begin{center}
    \begin{small}
      \begin{sc}
        \setlength{\tabcolsep}{2pt}
        \begin{tabular*}{\columnwidth}{@{\extracolsep{\fill}}lccc}
          \toprule
          Method & Best & Final & Train @10k \\
          \midrule
          GD+Euler & 1.107 & 1.113 & 0.962 \\
          Polyak+Euler & 1.099 & 1.110 & 0.933 \\
          Nesterov+Euler & 1.101 & 1.112 & 0.953 \\
                    GD+Lie--Trotter & 1.106 & 1.114 & 0.985 \\
                    Polyak+Lie--Trotter & 1.083 & 1.096 & 0.899 \\
          Nesterov+Lie--Trotter & {\bf 1.078} & {\bf 1.090} & {\bf 0.896} \\
          \bottomrule
        \end{tabular*}
      \end{sc}
    \end{small}
  \end{center}
  \vskip -0.1in
\end{table}

\begin{figure*}[t]
  \centering
  \subfigure[\texttt{small}]{\includegraphics[width=0.48\textwidth]{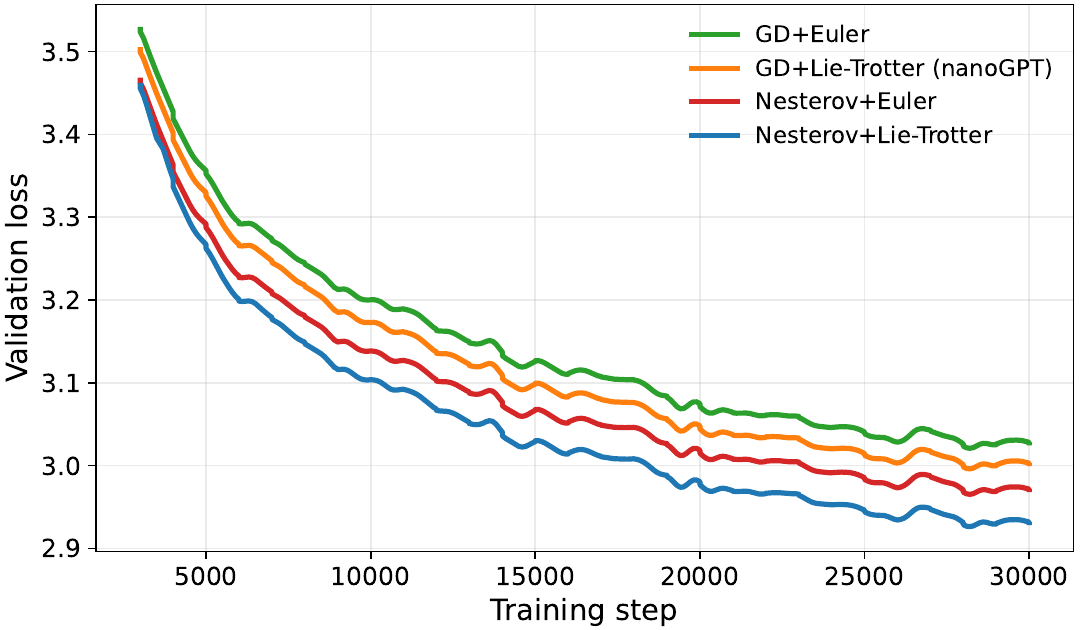}}
  \hfill
  \subfigure[\texttt{medium}]{\includegraphics[width=0.48\textwidth]{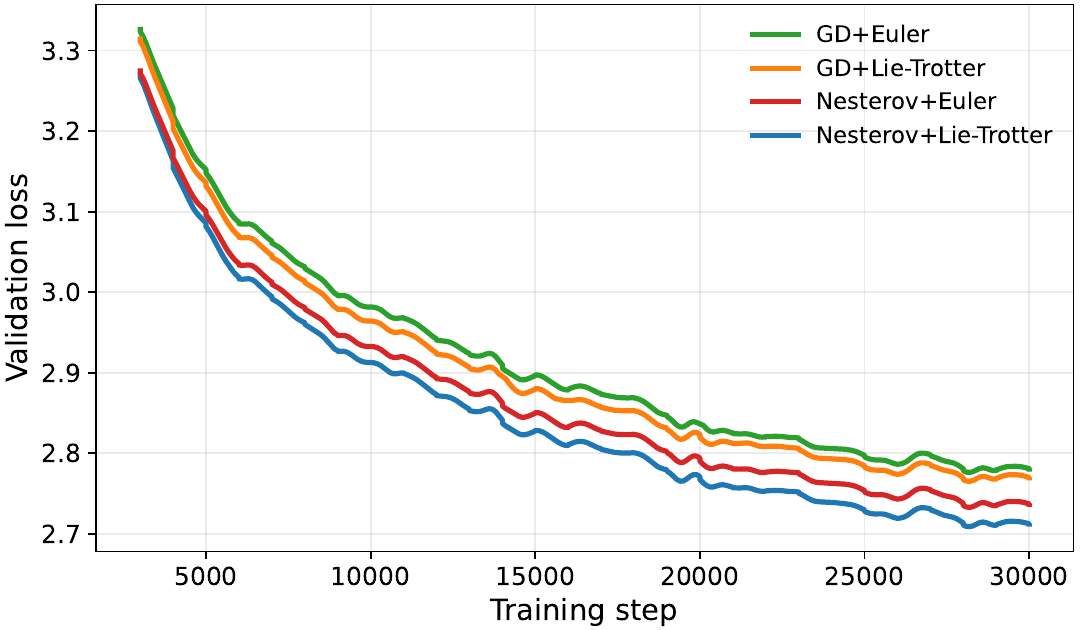}}
  \caption{OpenWebText: validation loss vs.\ training step for 12-layer and 24-layer models, excluding the first 3k warm-up steps and lightly Gaussian-smoothed ($\sigma=200$ steps) for readability. The ordering is consistent across training: GD+Euler is highest, Nesterov+Lie--Trotter is lowest, and GD+Lie--Trotter (nanoGPT) and Nesterov+Euler lie in between. See Figure~\ref{fig:nesterov-vs-polyak-openwebtext} for a direct Nesterov vs.\ Polyak comparison.}
  \label{fig:openwebtext-loss-curves}
\end{figure*}

\paragraph{OpenWebText ({\tt small} and {\tt medium}; 30k steps).}
Figure~\ref{fig:openwebtext-loss-curves} shows OpenWebText validation loss for \texttt{small} and \texttt{medium} models trained for 30k steps under identical data order and hyperparameters across methods.
After the 3k-step warmup, the ordering is stable for both sizes: GD+Euler is highest, Nesterov+Lie--Trotter is lowest, and GD+Lie--Trotter, Nesterov+Euler, and the Polyak variants lie in between.

\begin{table}[ht]
  \caption{OpenWebText ({\tt small} and {\tt medium}; 30k steps): next-token prediction validation losses (nats/token; lower is better). Best = min validation loss; Final = validation loss at 30k steps.}
  \label{tab:openwebtext-losses}
  \begin{center}
    \begin{small}
      \begin{sc}
        \setlength{\tabcolsep}{2pt}
        \begin{tabular*}{\columnwidth}{@{\extracolsep{\fill}}lcccc}
          \toprule
          & \multicolumn{2}{c}{{\tt small}} & \multicolumn{2}{c}{{\tt medium}} \\
          \cmidrule(lr){2-3} \cmidrule(lr){4-5}
          Method & Best & Final & Best & Final \\
          \midrule
          GD+Euler & 3.014 & 3.023 & 2.770 & 2.775 \\
          Polyak+Euler & 2.965 & 2.972 & 2.730 & 2.734 \\
          Nesterov+Euler & 2.959 & 2.966 & 2.726 & 2.731 \\
                    GD+Lie--Trotter & 2.990 & 2.998 & 2.758 & 2.764 \\
                    Polyak+Lie--Trotter & 2.925 & 2.931 & 2.705 & 2.710 \\
          Nesterov+Lie--Trotter & {\bf 2.920} & {\bf 2.926} & {\bf 2.702} & {\bf 2.707} \\
          \bottomrule
        \end{tabular*}
      \end{sc}
    \end{small}
  \end{center}
	  \vskip -0.1in
\end{table}

\subsection{Downstream tasks performance}
Table~\ref{tab:openwebtext-downstream} summarizes downstream multiple-choice
accuracy for the best-validation OpenWebText checkpoint of each method on
HellaSwag~\cite{zellers2019hellaswag} and
ARC-Easy~\cite{clark2018thinksolvedquestionanswering} using the Language Model
Evaluation Harness~\cite{eval-harness}.
We evaluate a fixed few-shot protocol (10-shot HellaSwag, 25-shot ARC-Easy;
plus 0-shot for reference) and report length-normalized accuracy
(\texttt{acc\_norm}, with standard errors).
We use the Language Model Evaluation Harness default few-shot seed
(\texttt{fewshot\_seed=1234}) and left-truncation for inputs exceeding the
1024-token context.

\begin{table*}[t]
  \caption{OpenWebText: length-normalized accuracy (\texttt{acc\_norm}; \%; higher is better) on HellaSwag (HS) and ARC-Easy (ARC) using the Language Model Evaluation Harness, evaluated at the best-validation checkpoint. Few-shot uses 10-shot HellaSwag and 25-shot ARC-Easy; 0-shot results are shown for reference. Errors are standard errors. Bold indicates the top-two methods in each column; ties at the reported precision are boldfaced (e.g., \texttt{small} ARC 0-shot has a tie for second at 41.1\%).}
  \label{tab:openwebtext-downstream}
  \begin{center}
    \begin{small}
      \begin{sc}
        \setlength{\tabcolsep}{2pt}
        \begin{tabular*}{\textwidth}{@{\extracolsep{\fill}}lcccccccc}
          \toprule
          & \multicolumn{4}{c}{\textnormal{\texttt{small}}} & \multicolumn{4}{c}{\textnormal{\texttt{medium}}}   \\
          \cmidrule(lr){2-5} \cmidrule(lr){6-9}
          & \multicolumn{2}{c}{10/25-shot} & \multicolumn{2}{c}{0-shot} & \multicolumn{2}{c}{10/25-shot} & \multicolumn{2}{c}{0-shot} \\
          \cmidrule(lr){2-3} \cmidrule(lr){4-5} \cmidrule(lr){6-7} \cmidrule(lr){8-9}
          Method & HS & ARC & HS & ARC & HS & ARC & HS & ARC \\
          \midrule
          GD+Euler & 29.9$\pm$0.5 & 39.8$\pm$1.0 & 30.2$\pm$0.5 & 38.5$\pm$1.0 & 34.7$\pm$0.5 & 46.0$\pm$1.0 & 34.4$\pm$0.5 & 40.5$\pm$1.0 \\
          Polyak+Euler & 30.9$\pm$0.5 & 41.3$\pm$1.0 & 30.5$\pm$0.5 & 39.8$\pm$1.0 & 36.3$\pm$0.5 & 48.2$\pm$1.0 & 36.1$\pm$0.5 & 43.4$\pm$1.0 \\
          Nesterov+Euler & 31.2$\pm$0.5 & 41.6$\pm$1.0 & 30.5$\pm$0.5 & {\bf 41.2$\pm$1.0} & 36.3$\pm$0.5 & 48.0$\pm$1.0 & 35.4$\pm$0.5 & 43.5$\pm$1.0 \\
                    GD+Lie--Trotter & 30.0$\pm$0.5 & 40.5$\pm$1.0 & 29.7$\pm$0.5 & 39.4$\pm$1.0 & 35.5$\pm$0.5 & 45.4$\pm$1.0 & 35.2$\pm$0.5 & 42.6$\pm$1.0 \\
                     Polyak+Lie--Trotter & {\bf 31.6$\pm$0.5} & {\bf 42.5$\pm$1.0} & {\bf 31.2$\pm$0.5} & {\bf 41.1$\pm$1.0} & {\bf 37.1$\pm$0.5} & {\bf 49.0$\pm$1.0} & {\bf 36.8$\pm$0.5} & {\bf 44.0$\pm$1.0} \\
          Nesterov+Lie--Trotter & {\bf 31.8$\pm$0.5} & {\bf 42.6$\pm$1.0} & {\bf 31.2$\pm$0.5} & {\bf 41.1$\pm$1.0} & {\bf 36.8$\pm$0.5} & {\bf 48.5$\pm$1.0} & {\bf 36.5$\pm$0.5} & {\bf 45.7$\pm$1.0} \\
          \bottomrule
        \end{tabular*}
      \end{sc}
    \end{small}
  \end{center}
  \vskip -0.1in
\end{table*}

The OpenWebText validation-loss ordering (Figure~\ref{fig:openwebtext-loss-curves}, Table~\ref{tab:openwebtext-losses}) largely holds for downstream \texttt{acc\_norm} (Table~\ref{tab:openwebtext-downstream}).
For example, Nesterov+Lie--Trotter improves few-shot HellaSwag over GD+Lie--Trotter from 30.0\% to 31.8\% (\texttt{small}) and from 35.5\% to 36.8\% (\texttt{medium}).
For ARC-Easy on \texttt{small}, the 25-shot gain is clearer than 0-shot (40.5\% to 42.6\% vs.\ 39.4\% to 41.1\%), motivating our emphasis on few-shot results.

\subsection{Discussion}


    To contextualize absolute OpenWebText losses, we compare against nanoGPT baselines for evaluating GPT-2 checkpoints on its OpenWebText preprocessing and validation split~\cite{karpathy2022nanogpt}. In that setup, GPT-2 (124M) attains a validation loss of $3.12$, while GPT-2 medium (350M) attains $2.84$. Training directly on OpenWebText, our {\tt small} model reaches $2.92$ (Nesterov+Lie--Trotter), below the GPT-2 (124M) checkpoint ($3.12$). At the medium scale, our {\tt medium} model reaches $2.70$, below the GPT-2 medium checkpoint ($2.84$). For reference, nanoGPT can reach $2.85$ at the {\tt small} scale, but this uses a much larger budget (600k steps / 294.9B tokens vs.\ our 30k steps / 14.75B tokens). Finally, absolute losses are not directly comparable due to differences in preprocessing, optimizers, and hyperparameters.

As a lightweight sanity check on our Language Model Evaluation Harness, we compare against GPT-2 ({\tt small}) under the same evaluation protocol (HellaSwag 10-shot; ARC-Easy 0-shot; \texttt{acc\_norm}). GPT-2 ({\tt small}) achieves 31.53\% on HellaSwag and 39.48\% on ARC-Easy~\cite{openllm_gpt2_details,olm_gpt2_latest_modelcard}, which is broadly consistent with our {\tt small} OpenWebText checkpoints (HS 10-shot: 29.9--31.8\%; ARC-Easy 0-shot: 38.5--41.2\%; Table~\ref{tab:openwebtext-downstream}).

We observe that Lie--Trotter splitting consistently outperforms Euler discretization. This superiority is less pronounced for OWT than for TinyStories. Our results also seem to indicate that it is the combination of Nesterov with Lie-Trotter that yields maximal improvement over the baseline.

\section{Extensions}
\label{sec:extensions}

Once attention and MLP sublayers are fixed as learned first-order oracles, alternative transformer
architectures arise by changing the outer update rule used to propagate token representations across
depth, together with the splitting used to compose oracle calls.
This yields a family of architectures corresponding to different discrete optimization schemes
applied to the same implicit objective.

As a simple illustration of this modularity, we consider Polyak's heavy-ball method~\cite{Polyak1964}, which replaces
Nesterov's lookahead-based acceleration with momentum evaluated at the current iterate.
The resulting architecture is obtained by applying a heavy-ball update to the
same attention and MLP oracles, with gradients evaluated at $X_t$ rather than at a lookahead point
$X^{\mathrm{in}}_t$.
Equivalently, it corresponds to the $\mu_t = 0$ (no-lookahead) specialization of
\textsc{YuriiFormer}, mirroring the classical relationship between Nesterov acceleration and Polyak
momentum.

Tables~\ref{tab:tinystories-losses} and~\ref{tab:openwebtext-losses} report losses for gradient descent,
Nesterov, and Polyak variants, each combined with either Euler or Lie--Trotter splitting.



\paragraph{Polyak with Euler discretization.}
The heavy-ball update in this context takes the form
\[
\begin{aligned}
V_{t+1} &= \beta_t V_t + \gamma_t {\sf Attn}_t(X_t) + \gamma_t {\sf MLP}_t(X_t),\\
X_{t+1} &= X_t + V_{t+1}.
\end{aligned}
\]
where $\beta_t\in(0,1)$ is a momentum parameter and $\gamma_t>0$ is a step size (both learned).

\paragraph{Polyak with Lie--Trotter splitting.}
Alternatively, the heavy-ball template can be combined with Lie--Trotter splitting to obtain a standard
sequential composition of attention and MLP layers:
\[
\begin{aligned}
V_{t+\frac12} &= \beta_t V_t + \gamma_t {\sf Attn}_t(X_t),\\
X_{t+\frac12} &= X_t + V_{t+\frac12},\\
V_{t+1} &= \beta_{t+\frac12} V_{t+\frac12} + \gamma_{t+\frac12} {\sf MLP}_t(X_{t+\frac12}),\\
X_{t+1} &= X_{t+\frac12} + V_{t+1}.
\end{aligned}
\]
with separate learned scalars for the two substeps.



\paragraph{Empirical comparison.}
On TinyStories and OpenWebText, Polyak achieves validation losses comparable to the Nesterov variants and improves over the GD baselines (Tables~\ref{tab:tinystories-losses} and~\ref{tab:openwebtext-losses}).
Compared to Polyak, Nesterov's lookahead yields a small additional loss improvement at the same compute and parameter budget (no extra attention/MLP oracle calls).
See Figure~\ref{fig:nesterov-vs-polyak-openwebtext} in the appendix for a direct Nesterov vs. Polyak comparison on OpenWebText.

\paragraph{Additional architectures.}
We also evaluate a few additional update-rule variants, including Verlet and IMEX.
Details and results are in Appendix~\ref{sec:additional-experiments} (Table~\ref{tab:tinystories-comprehensive}).

On TinyStories (\texttt{small}), under a matched budget of one attention and one MLP oracle call per block, Nesterov+Lie--Trotter is tied for best among the one-attention/one-MLP variants.
Variants that use additional attention/MLP oracle calls per block can achieve lower losses at higher compute.

\section{Conclusion}

We presented a variational and algorithmic framework that interprets transformer blocks as discrete optimization algorithms acting on token configurations.
Within this framework, attention and MLP sublayers play the role of learned first-order oracles for interaction and potential energies, and standard GPT-style transformers emerge naturally as implementations of gradient descent on a composite objective via Lie--Trotter splitting.
Framing transformers in this way shifts architectural design from heuristic modification to the principled selection of optimization templates and splitting schemes.

As a concrete instantiation of this perspective, we introduced momentum-based transformer architectures obtained by replacing gradient descent with classical accelerated methods while preserving the same attention and MLP oracle structure.
In particular, Both the Nesterov and Polyak variants of {\sc YuriiFormer} embed momentum directly at the representation level without increasing the number of attention or MLP evaluations per block.
Across language modeling benchmarks, both architectures consistently outperform vanilla GPT-style transformers of the same size and training budget, with Nesterov-style acceleration yielding the strongest and most stable improvements in validation loss and consistently improved downstream accuracy.

Taken together, these results demonstrate that viewing transformers as optimization algorithms is not only conceptually unifying but also practically useful.
More broadly, this framework opens the door to systematically importing ideas from numerical optimization and splitting methods to guide the design of new transformer architectures.

\section{Limitations}

This work is primarily concerned with architectural design rather than establishing formal convergence or optimality guarantees.
Although the optimization viewpoint provides a principled template for constructing transformer blocks, the implicit objectives induced by learned attention and MLP oracles are nonconvex, layer-dependent, and strongly shaped by preconditioning, placing them outside the scope of classical optimization theory.
Accordingly, we view optimization algorithms as design analogies rather than literal solvers whose theoretical guarantees are expected to transfer.

Our empirical evaluation focuses on small- and medium-scale language models under controlled training budgets, enabling clean comparisons between update rules.
While the observed improvements are consistent across datasets, model sizes, and tasks, evaluating these architectures at larger scales and in longer-context regimes remains an important direction for future work.

\section*{Impact Statement}

This work advances the theoretical and practical understanding of transformer architectures by providing a unified optimization-based framework for their design.
By connecting modern transformer blocks to classical optimization methods, it offers a systematic approach for developing new architectures that improve efficiency and performance.
The proposed methods are broadly applicable to sequence modeling tasks and can be readily integrated into existing transformer-based systems.
We hope this perspective will stimulate further research at the intersection of optimization, numerical analysis, and deep learning, and contribute to the continued development of reliable and effective machine learning models.

\bibliography{references}
\bibliographystyle{icml2026}

\appendix

\section{Experimental details and reproducibility}
\label{app:exp-setup}

This appendix records experimental setup details for Section~\ref{sec:experiments} that are omitted from the main text,
including parameter-counting conventions, dataset preprocessing, optimizer parameter groups, and evaluation/checkpointing.

\subsection{Models (additional details)}
All models use dropout $0$ and no bias terms.
For variants that maintain an additional velocity state, we initialize $v_0$ using token and positional embedding tables separate from the main token and positional embeddings.

\paragraph{Models and parameter counts.}
We follow a common GPT non-positional counting convention: letting $V$ be the vocabulary size and $d$ the embedding dimension, we exclude the learned positional embedding table (which scales with context length) but include the token embedding matrix (size $V\times d$) since it is weight-tied to the output projection used for next-token logits.

Under this convention, the 12L/12H/768d and 24L/16H/1024d GD baselines (GD+Lie--Trotter (nanoGPT) and GD+Euler) have 123.6M and 353.6M non-positional parameters, respectively.
The Nesterov-accelerated variants (Nesterov+Euler and Nesterov+Lie--Trotter) keep the same token embedding and Transformer weight matrices as the GD baselines.
Under this convention, they therefore have the same non-positional parameter counts as above (123.6M for 12L/12H/768d and 353.6M for 24L/16H/1024d), up to a negligible overhead from the update rule.

Additional learned parameters include a velocity LayerNorm ($O(Ld)$, with $L$ layers) and $O(L)$ learned scalars.
These update-rule scalars are learned end-to-end and constrained via reparameterization: scalars constrained to $(0,1)$ are stored as logits and mapped through a sigmoid, while positive scalars are mapped via \texttt{softplus}.
Momentum variants also learn separate $v_0$ embedding tables used to initialize the velocity: a token embedding table (size $Vd$, adding 38.6M parameters for $d{=}768$ and 51.5M for $d{=}1024$), and a positional embedding table (size $Td$), which is excluded from non-positional parameter counts by the convention above.
\subsection{Datasets and preprocessing}

We tokenize with the GPT-2 byte-level BPE tokenizer (\texttt{tiktoken}; GPT-2 BPE with 50{,}257 tokens, with model vocabulary padded to $V{=}50{,}304$ as in nanoGPT) and train on contiguous sequences of length $T{=}1024$ with next-token prediction targets.
We append an end-of-text token between documents/stories.
We train in epochs over non-overlapping $T$-token blocks: each epoch visits every block exactly once, and between epochs we shift block boundaries by a seeded offset and reshuffle the block order (so the full sequence of batches is deterministic given the seed).

We use TinyStories and OpenWebText (introduced in Section~\ref{sec:experiments}).
After tokenization, our TinyStories split contains 471.6M training tokens and 2.37M validation tokens.

Our processed OpenWebText contains 8{,}013{,}769 documents and 9.05B tokens.
We deterministically assign documents to train/validation splits with a fixed split targeting 95/5 based on the seed and document index, yielding 7{,}630{,}278 training documents (8.62B tokens) and 383{,}491 validation documents (432M tokens).

\subsection{Optimization and hyperparameters}

\paragraph{Optimizer (parameter groups).}
Muon updates non-embedding matrix-valued parameters (2D+ weight matrices) with zero weight decay.
AdamW updates embeddings (including the weight-tied token embedding/output matrix, and separate velocity embeddings when present) with weight decay 0.1.
AdamW uses zero weight decay for normalization parameters and learned scalar update-rule parameters (when present); the scalar group uses a $5\times$ learning-rate multiplier.

\paragraph{Evaluation.}
We compute validation loss on the validation split every 100 optimizer steps.
Each evaluation averages over 160 validation batches (4{,}915{,}200 tokens).

\paragraph{Checkpointing.}
We save the best checkpoint whenever validation loss improves, and save a final checkpoint at the end of training.

\section{Additional Experiments}
\label{sec:additional-experiments}

We compare update-rule variants under fixed model size, optimizer, data order, and number of optimizer steps,
focusing on TinyStories as the primary benchmark. Some variants use extra attention/MLP oracle calls per block,
so compute per step is higher where noted.

\subsection{Polyak's heavy ball}
\label{app:borisformer}

\begin{figure}[t]
  \centering
  \includegraphics[width=0.23\textwidth,height=0.35\textheight,keepaspectratio]{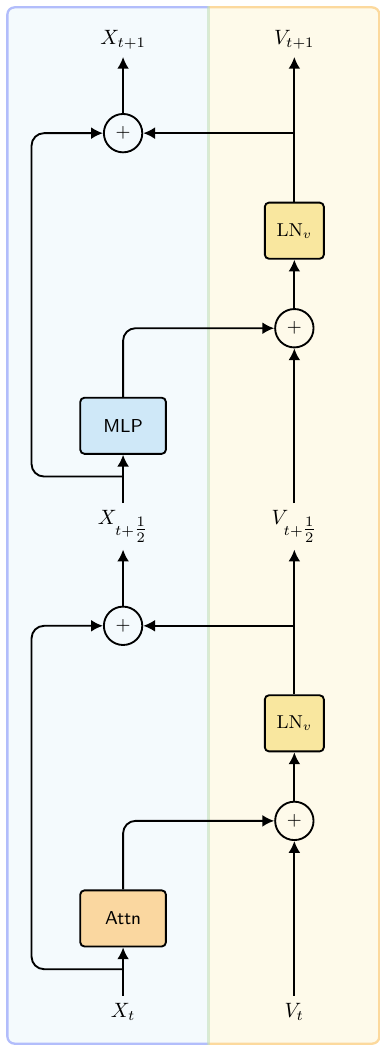}
  \caption{Polyak variant of {\sc YuriiFormer} with Lie--Trotter splitting (no lookahead), with velocity LayerNorm $\mathrm{LN}_v$ applied after each velocity update.}
  \label{fig:borisformer-arch}
\end{figure}

\begin{figure*}[t]
  \centering
  \subfigure[\texttt{small}]{\includegraphics[width=0.48\textwidth]{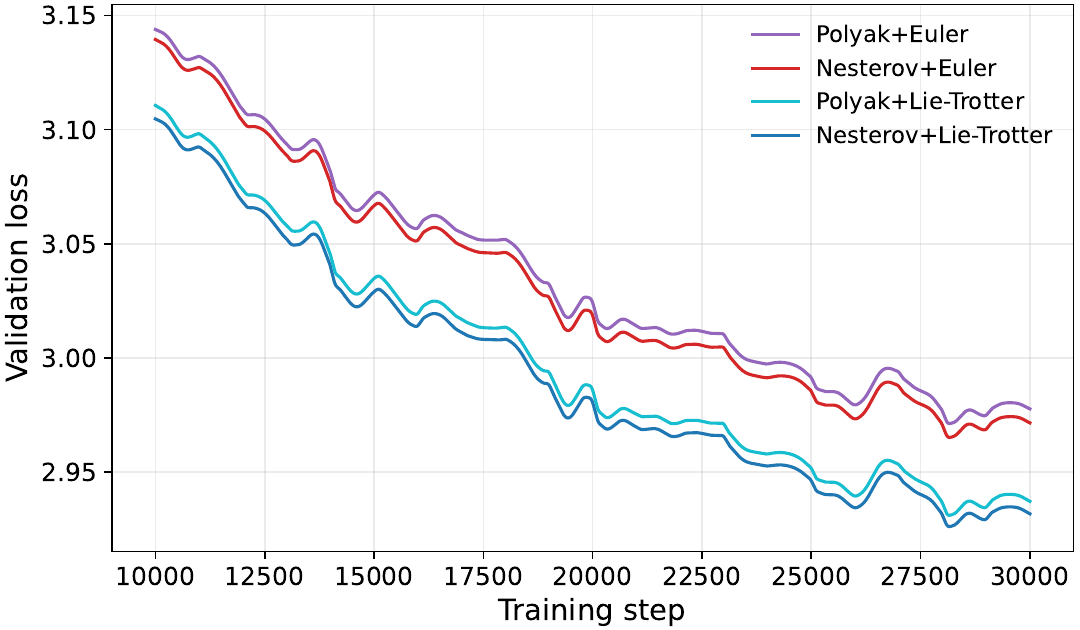}}
  \hfill
  \subfigure[\texttt{medium}]{\includegraphics[width=0.48\textwidth]{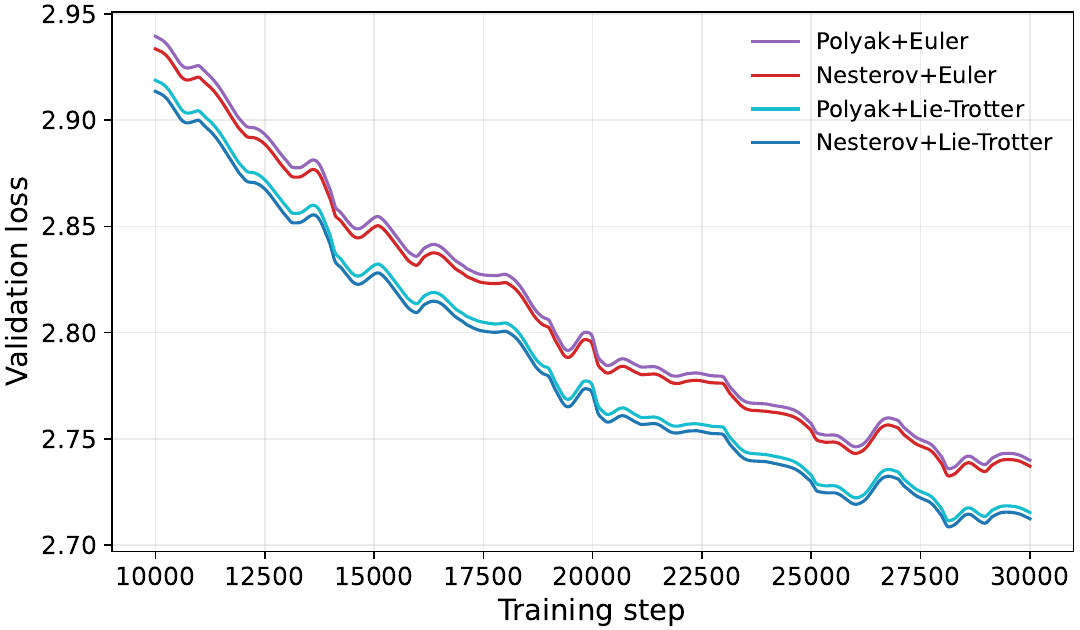}}
  \caption{OpenWebText: Nesterov vs. Polyak variants \textsc{YuriiFormer} validation loss for \texttt{small} and \texttt{medium} models (steps $\ge 10$k). Curves are nearly identical across sizes, with a small but consistent advantage for Nesterov.}
  \label{fig:nesterov-vs-polyak-openwebtext}
\end{figure*}

Figure~\ref{fig:borisformer-arch} shows the Polyak block used in our experiments; see Section~\ref{sec:extensions} for the update equations.
Figure~\ref{fig:nesterov-vs-polyak-openwebtext} compares the Nesterov and Polyak variants and \textsc{YuriiFormer} on OpenWebText.

\subsection{Additional update-rule variants}
\label{app:update-rule-variants}

Throughout this subsection we use the main-text notation: $X_t$ denotes the token configuration at depth $t$, and momentum methods maintain an auxiliary velocity $V_t$.
We write the attention and MLP updates as ${\sf Attn}_t(\cdot)$ and ${\sf MLP}_t(\cdot)$; split variants apply attention then MLP (Lie--Trotter).

\paragraph{Notation and conventions.}
Throughout, ${\sf Attn}_t(\cdot)$ and ${\sf MLP}_t(\cdot)$ denote the residual directions produced by the attention and MLP sublayers at depth $t$
(including learned linear maps and normalizations).
We use superscripts ${\mathrm{in}}$ to denote lookahead evaluation points (e.g., $X_t^{\mathrm{in}}=X_t+\mu_t V_t$); indices such as $t+\frac12$ denote intermediate quantities within a layer update.
Unless stated otherwise, all scalar coefficients in these updates are learned.

\paragraph{IMEX (implicit/explicit split).}
IMEX (implicit-explicit) schemes split the composite update (the sum of the attention and MLP updates) into
an explicit map and an implicit map; the explicit map is evaluated at a lookahead of the current state, while the implicit
map is defined at the new state and is approximated here with $k$ unrolled refinements~\cite{ascher1995imex,pareschi2005imex}.

We evaluate both Attn-explicit/MLP-implicit and MLP-explicit/Attn-implicit orderings, with two variants: standard IMEX
(LN$_v$ once at the end) and IMEX+LN$_v$ (LN$_v$ after every velocity update). For clarity we write the
Attn$\rightarrow$MLP ordering; MLP$\rightarrow$Attn is identical with Attn/MLP swapped.
Because these two IMEX variants differ only in where velocity normalization is applied, we write $\mathrm{LN}_v$ explicitly in the IMEX updates (and omit it from the other variant equations below for readability).
Define the explicit attention step
\[
X_{t+\frac12}=X_t+\mu_t V_t,\qquad V_{t+\frac12}=\beta_t V_t+\gamma_t\,{\sf Attn}_t(X_{t+\frac12}).
\]
\emph{Implicit MLP with $k$ unrolled refinements.} Initialize
\[
V_{t+1}^{(0)}=V_{t+\frac12},\qquad X_{t+\frac12}^{(0)}=X_t+\mu_{t+\frac12}V_{t+1}^{(0)},
\]
and for $j=1,\ldots,k$ iterate
\[
\begin{aligned}
V_{t+1}^{(j)} &= \beta_t V_{t+\frac12}+\gamma_{t+\frac12}\,{\sf MLP}_t\!\big(X_{t+\frac12}^{(j-1)}\big),\\
X_{t+\frac12}^{(j)} &= X_t+\mu_{t+\frac12}V_{t+1}^{(j)}.
\end{aligned}
\]
Finally, apply a single velocity normalization and update the state:
\[
V_{t+1}=\mathrm{LN}_v\!\big(V_{t+1}^{(k)}\big),\qquad X_{t+1}=X_t+V_{t+1}.
\]
For $k=1$ (one implicit refinement; larger $k$ increases compute time), the full forward update is
\[
\begin{aligned}
X_{t+\frac12} &= X_t+\mu_t V_t,\\
V_{t+\frac12} &= \beta_t V_t+\gamma_t\,{\sf Attn}_t(X_{t+\frac12}),\\
X_{t+\frac12}^{\mathrm{in}} &= X_t+\mu_{t+\frac12}V_{t+\frac12},\\
V_{t+1} &= \mathrm{LN}_v\!\Big(\beta_t V_{t+\frac12}+\gamma_{t+\frac12}\,{\sf MLP}_t(X_{t+\frac12}^{\mathrm{in}})\Big),\\
X_{t+1} &= X_t+V_{t+1}.
\end{aligned}
\]
Here $\beta_t$ is shared across substeps, $\gamma_t$ and $\gamma_{t+\frac12}$ are the explicit/implicit step sizes, and $\mu_t,\mu_{t+\frac12}$ are the corresponding lookahead coefficients.
IMEX+LN$_v$ differs only in applying $\mathrm{LN}_v$ after the explicit step and after each implicit
refinement (i.e., after every velocity update).
We test both orderings (Attn$\rightarrow$MLP and MLP$\rightarrow$Attn), both variants (standard IMEX and IMEX+LN$_v$),
and $k\in\{1,2\}$.

Compute per block is 1 Attn + $k$ MLP (Attn$\rightarrow$MLP) or 1 MLP + $k$ Attn (MLP$\rightarrow$Attn).
For $k=1$ we obtain the minimal IMEX scheme (baseline compute), but it still differs from the Lie--Trotter block because the implicit map uses the lookahead $X_t+\mu_{t+\frac12}V_{t+1}^{(0)}$.

\makeatletter
\ifdefined\saved@dblfptop\else\newlength{\saved@dblfptop}\fi
\setlength{\saved@dblfptop}{\@dblfptop}
\setlength{\@dblfptop}{0pt plus 0fil}
\makeatother
\begin{table*}[!t]
  \caption{TinyStories (\texttt{small}; 10k steps): comprehensive next-token prediction losses (nats/token; lower is better). Metrics: Best (minimum validation loss), Final (validation loss at 10k steps), Train @10k (training loss at 10k steps). Bold indicates the best value within each panel (standard vs non-standard compute) for each metric; ties at the reported precision are boldfaced.}
  \label{tab:tinystories-comprehensive}
  \begin{center}
    \begin{small}
      \begin{sc}
        \setlength{\tabcolsep}{2pt}
        \begin{tabular*}{\textwidth}{@{\extracolsep{\fill}}lccc|lccc}
          \toprule
          \multicolumn{4}{c}{\textbf{Standard compute (1 Attn + 1 MLP)}} & \multicolumn{4}{c}{\textbf{Non-standard compute}} \\
          \cmidrule(lr){1-4} \cmidrule(lr){5-8}
          Method & Best & Final & Train @10k & Method & Best & Final & Train @10k \\
          \midrule
          GD+Euler & 1.107 & 1.113 & 0.962 & PRK-Verlet (MAM) & 1.077 & 1.090 & 0.886 \\
          GD+Lie--Trotter & 1.106 & 1.114 & 0.985 & PRK-Verlet (AMA) & 1.074 & 1.088 & {\bf 0.874} \\
          Nesterov+Euler & 1.101 & 1.112 & 0.953 & IMEX $k$=2 (MAM) & 1.081 & 1.096 & 0.921 \\
          Nesterov+Lie--Trotter & {\bf 1.078} & {\bf 1.090} & 0.896 & IMEX $k$=2 (AMA) & 1.078 & 1.088 & 0.921 \\
          Polyak+Euler & 1.099 & 1.110 & 0.933 & IMEX+LN$_v$ $k$=2 (MAM) & 1.097 & 1.104 & 0.966 \\
          Polyak+Lie--Trotter & 1.083 & 1.096 & 0.899 & IMEX+LN$_v$ $k$=2 (AMA) & {\bf 1.070} & {\bf 1.084} & 0.884 \\
          Hamiltonian (symplectic Euler) & 1.080 & 1.092 & 0.916 & & & & \\
          IMEX $k$=1 (MAM) & 1.083 & 1.095 & 0.912 & & & & \\
          IMEX $k$=1 (AMA) & 1.086 & 1.097 & 0.929 & & & & \\
          IMEX+LN$_v$ $k$=1 (MAM) & 1.112 & 1.123 & 1.008 & & & & \\
          IMEX+LN$_v$ $k$=1 (AMA) & {\bf 1.078} & 1.091 & {\bf 0.892} & & & & \\
          \bottomrule
        \end{tabular*}
      \end{sc}
    \end{small}
  \end{center}
  \vskip -0.1in
\end{table*}
\makeatletter\setlength{\@dblfptop}{\saved@dblfptop}\makeatother

\paragraph{PRK-Verlet (Strang splitting).}
We adopt the velocity-Verlet scheme of Verlet~\cite{verlet1967computer}, applying a symmetric half-step/full-step/
half-step composition to the split attention/MLP directions. This is the classical Strang symmetric composition,
yielding a second-order, time-reversible update for the split dynamics~\cite{strang1968construction}.

Starting from the composite objective $\cE+\cF$ with update maps ${\sf Attn}_t$ and ${\sf MLP}_t$, PRK-Verlet
uses the Attn$\rightarrow$MLP$\rightarrow$Attn symmetric split update (Strang): a half-step in attention, a full step in the MLP,
and a second half-step in attention. By ``half-step'' we mean scaling the attention step size by $1/2$ while keeping the same damping.
For the Attn$\rightarrow$MLP$\rightarrow$Attn ordering, the updates are:
\par\noindent\textbf{Attn half-step.}
\[
\begin{aligned}
X_t^{\mathrm{in}} &= X_t+\mu_t V_t,\\
V_{t+\frac12} &= \beta_t V_t+\tfrac12\gamma_t\,{\sf Attn}_t(X_t^{\mathrm{in}}),\\
X_{t+\frac12} &= X_t+V_{t+\frac12}.
\end{aligned}
\]
\textbf{MLP full step.}
\[
\begin{aligned}
X_{t+\frac12}^{\mathrm{in}} &= X_{t+\frac12}+\mu_{t+\frac12}V_{t+\frac12},\\
V_{t+\frac12}^\ast &= \beta_t V_{t+\frac12}+\gamma_{t+\frac12}\,{\sf MLP}_t(X_{t+\frac12}^{\mathrm{in}}),\\
X_{t+\frac12}^\ast &= X_{t+\frac12}+V_{t+\frac12}^\ast.
\end{aligned}
\]
\textbf{Attn half-step.}
\[
\begin{aligned}
X_{t+\frac12}^{\mathrm{in},\ast} &= X_{t+\frac12}^\ast+\mu_t V_{t+\frac12}^\ast,\\
V_{t+1} &= \beta_t V_{t+\frac12}^\ast+\tfrac12\gamma_t\,{\sf Attn}_t(X_{t+\frac12}^{\mathrm{in},\ast}),\\
X_{t+1} &= X_{t+\frac12}^\ast+V_{t+1}.
\end{aligned}
\]
We test both Attn$\rightarrow$MLP$\rightarrow$Attn and MLP$\rightarrow$Attn$\rightarrow$MLP orderings; compute per block is 2 Attn + 1 MLP for Attn$\rightarrow$MLP$\rightarrow$Attn and 2 MLP + 1 Attn for MLP$\rightarrow$Attn$\rightarrow$MLP (1.5$\times$ in oracle calls; actual FLOPs depend on ordering).
In our implementation, $\mathrm{LN}_v$ is applied after each velocity update (after each half/full step).

\paragraph{Hamiltonian (symplectic Euler).}
Motivated by symplectic-Euler--style updates for partitioned Hamiltonian systems~\cite{hairer2006gni}, we evolve a token state $X_t$ and momentum $V_t$ with an Attn-kick $\rightarrow$ drift $\rightarrow$ MLP-kick update. Each kick evaluates its map at a lookahead $X+\mu V$ and updates $V$ with damping multiplier $\beta$, while the drift updates $X$ by $\delta_t V$; the lookahead is recomputed after the drift. In this variant, the kicks subtract the residual directions (as implemented).
The update is
\[
\begin{aligned}
\textbf{Attn-kick:}\quad
X_t^{\mathrm{in}} &= X_t+\mu_t V_t,\\
V_{t+\frac12} &= \beta_t V_t - \gamma_t\,{\sf Attn}_t(X_t^{\mathrm{in}}),\\
\textbf{Drift:}\quad
X_{t+\frac12} &= X_t + \delta_t\,V_{t+\frac12},\\
\textbf{MLP-kick:}\quad
X_{t+\frac12}^{\mathrm{in}} &= X_{t+\frac12}+\mu_t V_{t+\frac12},\\
V_{t+1} &= \beta_t V_{t+\frac12} - \gamma_{t+\frac12}\,{\sf MLP}_t(X_{t+\frac12}^{\mathrm{in}}),\\
X_{t+1} &= X_{t+\frac12}.
\end{aligned}
\]
In our implementation, $\mathrm{LN}_v$ is applied after each momentum update.

\subsection{Comprehensive TinyStories Results}
\label{app:tinystories-full-results}

Table~\ref{tab:tinystories-comprehensive} presents loss metrics for all 17~update-rule variants evaluated on TinyStories
(12L/12H/768d, 10k steps). The left panel groups standard-compute variants (1 Attn + 1 MLP per block), while
the right panel lists non-standard-compute variants (extra Attn/MLP oracle calls). All variants
use identical model size, optimizer settings, data order, and step budget; non-standard methods incur
additional compute per step.

\end{document}